\documentclass[12pt]{article}
\usepackage{graphicx}
\usepackage{apalike}
\usepackage{amsmath}

\usepackage{algorithm,algorithmic}
\usepackage{epsfig}
\begin{document}
\renewcommand{\algorithmicrequire} {\textbf{\textsc{Entr\'ee}}}
\renewcommand{\algorithmicensure}  {\textbf{\textsc{Sorti}}}
\renewcommand{\algorithmicif}  {\textbf{\textsc{Si}}}
\renewcommand{\algorithmicthen}  {\textbf{\textsc{Alors}}}
\renewcommand{\algorithmicelse}  {\textbf{\textsc{Sinon}}}

\begin{center}
{\Large
	{\sc
		 Classification Recouvrante Bas\'ee sur les M\'ethodes \`a Noyau
	}
}
\bigskip

 Chiheb-Eddine Ben N'Cir \& Nadia Essoussi

\medskip
{\it

LARODEC,Institut Sup\'erieur de Gestion de Tunis, Universit\'e de Tunis
chiheb.benncir@gmail.com, nadia.essoussi@isg.rnu.tn

}
\end{center}
\bigskip
\noindent

\begin{abstract} Overlapping clustering problem is an important learning issue in which clusters are not mutually exclusive and each object may belongs simultaneously to several clusters. This paper presents a kernel based method that produces overlapping clusters on a high feature space using mercer kernel techniques to improve separability of input patterns. The proposed method, called OKM-K(Overlapping $k$-means based kernel method), extends OKM (Overlapping $k$-means) method to produce overlapping schemes. Experiments are performed on overlapping dataset and empirical results obtained with OKM-K outperform results obtained with OKM.
\begin{center}
    \textbf{R\'esum\'e}
\end{center}

Le probl\`eme de la classification recouvrante constitue un axe important de l'apprentissage automatique. Dans cet axe, les clusters ne sont pas mutuellement exclusifs et chaque objet peut appartenir simultan\'ement \`a plusieurs groupes appel\'es recouvrements. Cet article pr\'esente une m\'ethode \`a noyau permettant de produire des clusters non disjoints dans un espace de redescription fortement dimensionnel en utilisant les techniques de l'astuce de noyau pour am\'eliorer la s\'eparabilit\'e du mod\`ele de donn\'ees initial. La m\'ethode propos\'ee, OKM-K(Overlapping $k$-means based kernel method) \'etend la m\'ethode OKM(Overlapping $k$-means). Les exp\'erimentations sont effectu\'ees sur un ensemble de donn\'ees recouvrantes et les r\'esultats empiriques obtenus avec OKM-K sont meilleures que les r\'esultats obtenus avec OKM.
\end {abstract}

\noindent \textbf{\small{mots cl\'es:}} \textsl{\small{Apprentissage et Classification, Data Mining, M\'ethodes \`a Noyau}}

\section{Introduction}
Cet article s'int\'eresse au domaine de la classification recouvrante qui consiste \`a assigner des objets dans des classes non disjointes appel\'ees recouvrements (Cleuziou , 2007). En effet, plusieurs probl\`emes r\'eels n\'ecessitent qu'un objet puisse appartenir \`a la fois \`a plusieurs partitions. Par exemple, en biologie, un g\`ene peut participer \`a plusieurs processus ; en recherche d'information, un document peut aborder plusieurs th\'ematiques ou appartenir \`a plusieurs genres diff\'erents ; en traitement du langage, un mot peut avoir plusieurs interpr\'etations.

 Plusieurs m\'ethodes ont \'et\'e propos\'ees pour r\'esoudre ce probl\`eme. Les premi\`eres m\'ethodes \'etendent les sch\'emas de classification floue o\`u un objet appartient \`a plusieurs classes avec diff\'erents degr\'es d'appartenance(Deodhar and Ghosh,$2006$). En fixant un seuil minimal sur ces degr\'es, les objets sont affect\'es \`a une ou \`a plusieurs classes. Ces m\'ethodes ne permettent pas de traiter tous les sch\'emas de recouvrement possibles. Des m\'ethodes plus r\'ecentes de classification recouvrante ont r\'esolu ce probl\`eme en d\'eterminant directement des recouvrements optimaux et non pas des partitions optimales. L'\'eventail de ces m\'ethodes comprend \`a minima des g\'en\'eralisations des m\'ethodes de r\'eallocation dynamique telle que la m\'ethode OKM propos\'ee par Cleuziou ~(2008) , des adaptations des m\'ethodes des m\'elanges de lois (Banerjee, 2005), (Heller and Ghahramani, 2007) et des m\'ethodes fond\'ees sur la th\'eorie des graphes pour produire des sch\'emas recouvrants (Fellows et al.,$2009$).

Dans cet article, nous nous int\'eressons \`a la construction des classes recouvrantes ainsi qu'\`a la d\'etermination des s\'eparations non sph\'eriques entre les recouvrements. Nous proposons une m\'ethode qui combine les avantages de la m\'ethode OKM pour la construction directe des recouvrements optimaux et les avantages de la m\'ethode kernel $k$-means(Camastra and Verri, $2005$) pour la d\'etermination des classes ayant des formes non sh\'eriques.

\section {OKM: Overlapping $k$-means}

La m\'ethode OKM \'etend la m\'ethode $k$-moyennes pour chercher des recouvrements optimaux plut\^ot que des partitions optimales. Etant donn\'e un ensemble d'objets \`a classifier $ X={\{x_i\}}_{i=1} ^{N}$ avec $x_i \in \Re^d$ et $N$ le nombre d'objets, il s'agit de d\'eterminer les $k$ recouvrements de telle sorte que la fonction objective suivante soit optimis\'ee :

\begin{equation}\label{eq1}
     J(\pi)=\displaystyle \sum_{x_i \in X}\|x_i-im(x_i)\|^2 .
\end{equation}

\noindent La notation $im(x_i)$ d\'esigne l'image de $x_i$ d\'efinie par la combinaison des centres des clusters auxquels $x_i$ appartient :
\begin{equation}\label{eq2}
  im(x_i)=\displaystyle \sum_{c \in A_i}m_c / |A_i| ,
\end{equation}

\noindent o\`u $A_i$ est l'ensemble des affectations aux diff\'erents clusters de l'objet $x_i$, c'est-\`a-dire les clusters auquels $x_i$ appartient et $m_c$ correspond au centre du cluster $c$.

Le crit\`ere $J$ de la fonction objective g\'en\'eralise le crit\`ere des moindres carr\'es utilis\'es dans la m\'ethode k-moyennes. Pour minimiser ce crit\`ere, deux \'etapes principales sont ex\'ecut\'ees it\'erativement tant que le crit\`ere $J$ n'est pas minimis\'e. La premi\`ere \'etape consiste \`a calculer les centres des clusters en utilisant la fonction $PROTOTYPE$ (Cleuziou, $2008$). La deuxi\`eme \'etape consiste \`a affecter chaque objet \`a une ou \`a plusieurs classes selon la fonction d'affectation $ASSIGN$. La convergence de la m\'ethode est caract\'eris\'ee par plusieurs crit\`eres \`a savoir le nombre d'it\'erations maximales et le seuil minimal d'am\'elioration de la fonction objective entre deux it\'erations.

La m\'ethode OKM ne permet pas de d\'eterminer les classes de formes concentriques et les classes de formes non sph\'eriques. Pour r\'esoudre ce probl\`eme, nous proposons d'\'etendre OKM en utilisant les m\'ethodes \`a noyau.

\section {OKM-K: Overlapping $k$-means based kernel method}

 Le crit\`ere d'erreur de cette m\'ethode, tel que d\'efini dans eq.($3$), est optimis\'e dans un espace fortement dimensionnel pour am\'eliorer la recherche des s\'eparations entre les clusters.

\begin{eqnarray}
J(\pi) & = & \displaystyle\sum_{x_i\in X} \|\phi(x_i)-im(\phi(x_i))\|^2, \label{eq3}
\end{eqnarray}

\noindent avec $\phi(x_i)$ la repr\'esentation de l'objet $x_i$ dans le nouvel espace. L'image $im(\phi(x_i))$ est aussi d\'efinie dans l'espace de redescription par:

\begin{equation}\label{eq4}
 im(\phi(x_i))= \displaystyle\frac{\displaystyle\sum_{c=1 }^k P_{ic}.m_c^\phi}{\displaystyle\sum_{c=1 }^k P_{ic}},
\end{equation}

\noindent avec $P_{ic} \in \{0,1\}$ une variable binaire indiquant l'appartenance de l'objet $i$ au cluster $c$, et $m_{c}^\phi$ le prototype du cluster $c$ dans l'espace de redescription. Le prototype d'un cluster est d\'efini par le centre de gravit\'e des objets qui appartiennent \`a ce cluster pond\'er\'es par le nombre de clusters auquels chaque objet appartient comme illustr\'e dans eq.($5$):	

\begin{equation}\label{eq5}
 m_c^\phi= \displaystyle\frac{\displaystyle\sum_{j=1 }^N P_{jc}.\phi(x_j).w_j}{\displaystyle W_c} ,
\end{equation}

\noindent avec $W_c$, la somme des poids des objets qui appartiennent au cluster $c$ d\'efini par $W_{c}=\displaystyle\sum_{j=1}^N P_{jc}.w_j $. La notation $w_j$ indique le poids unitaire assign\'e \`a l'object $j$ d\'efini par $w_j= 1\displaystyle/(\sum_{c=1}^k P_{jc})^2$. A partir de cette d\'efinition des prototypes des clusters, le crit\`ere d'erreur peut \^etre calcul\'e comme suit:
\begin{eqnarray}
J(\pi) &=& \displaystyle \sum_{x_i\in X} \|\phi(x_i)- \frac{1}{L_i}\displaystyle \sum_{c=1}^k P_{ic}.\frac{1}{W_c} \sum_{j=1}^N P_{jc}.w_j.\phi(x_j)\|^2 \nonumber \\
&=&\sum_{x_i\in X} \{\phi(x_i).\phi(x_i) - \frac{2} {L_i}\sum_{c=1}^k \sum_{j=1}^N P_{ic}.\frac{1}{W_c}.P_{jc}.w_j.\phi(x_i).\phi(x_j) + \nonumber \\
& & \quad \displaystyle \frac{1} {(L_i)^2} \sum_{c=1}^k \sum_{j=1}^N \sum_{t=1}^k \sum_{g=1}^N P_{ic}.\frac{1}{W_c}.P_{jc}.P_{it}.\frac{1}{W_t}.P_{gt}.w_j.w_g.\phi(x_j)\phi(x_g)\}, \label{eq6}
\end{eqnarray}

\noindent avec $L_i=\displaystyle \sum_{c=1}^k P_{ic}$. En rempla\c{c}ant chaque produit scalaire dans l'espace de redescription par la fonction de noyau, le crit\`ere $J$ peut \^etre d\'etermin\'e sans r\'eellement d\'efinir les repr\'esentations $\phi(x_i)$:

\begin{eqnarray}
 J(\pi)&=& \displaystyle\sum_{x_i\in X} \{ \displaystyle K_{ii} - \frac{2} {L_i}\sum_{c=1}^k \sum_{j=1}^N P_{ic}.\frac{1}{W_c}.P_{jc}.w_j.K_{ij} + \nonumber \\
 & & \quad \quad \quad \displaystyle \frac{1} {(L_i)^2} \sum_{c=1}^k \sum_{j=1}^N \sum_{t=1}^k \sum_{g=1}^N  P_{ic}.\frac{1}{W_c}.P_{jc}.P_{it}.\frac{1}{W_t}.P_{gt}.w_j.w_g.K_{jg}  \},\label{eq7}
\end{eqnarray}

avec $K_{ij}$ est la fonction de noyau repr\'esenant le produit scalaire entre $\phi(x_i)$ et $\phi(x_j)$. Pour optimiser le crit\`ere d'erreur, la m\'ethode OKM-K affecte \`a chaque it\'eration les repr\'esentations des objets $\phi(x_i)$ \`a un ou plusieurs clusters puis elle calcule de nouveau le crit\`ere $J$ de la fonction objective. Si ce crit\`ere s'am\'eliore d'une it\'eration \`a une autre, les objets sont r\'eaffect\'es aux clusters les plus proches jusqu'\`a l'optimisation de ce crit\`ere. Les conditions d'arr\^et sont le nombre maximal d'it\'eration et l'am\'elioration minimale dans la fonction objective d'une it\'eration \`a une autre.
\section{Exp\'erimentations}
Nous avons compar\'e l'efficacit\'e de la m\'ethode propos\'ee OKM-K par rapport \`a la m\'ethode OKM sur la la base de donn\'ees $EachMovie$ \footnote {http://www.grouplens.org/node/76.} qui contient des \'evaluations en lignes des internautes pour certains films. Si chaque genre de film est consid\'er\'e comme une classe contenant plusieurs films, alors cette base de donn\'ees contient naturellement des classes recouvrantes. Un film peut appartenir \`a plusieurs genres.
\begin{figure}[!h]
  \vspace{-0.2cm}
  \centering
   {\epsfig{file = 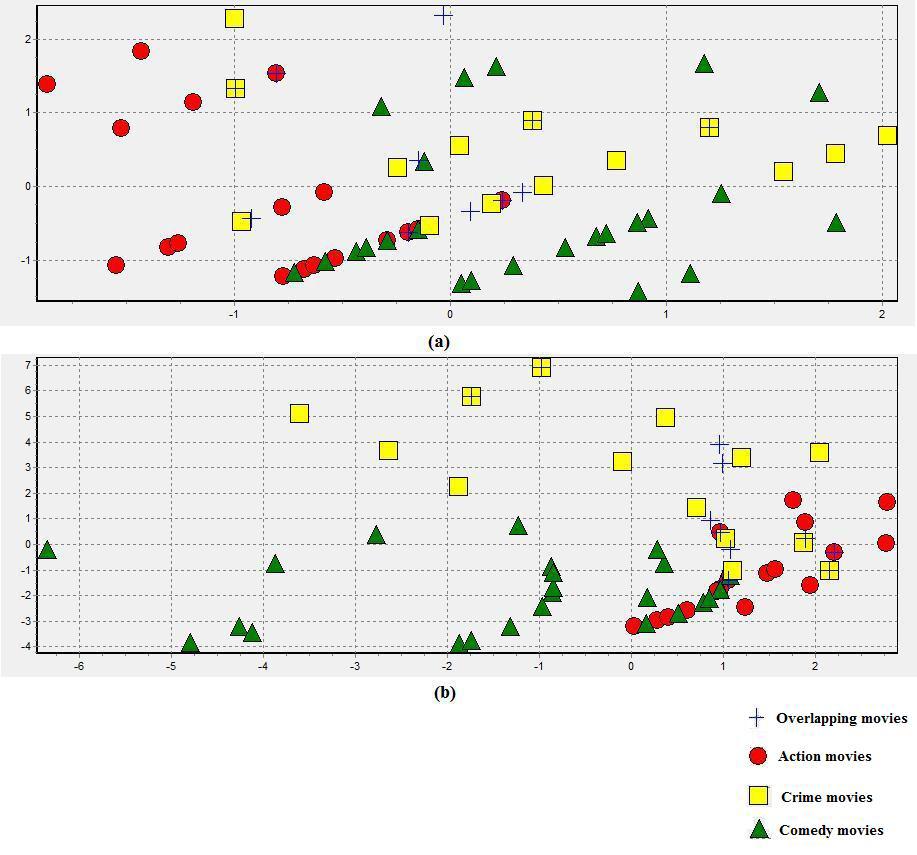, width = 8 cm, height= 8 cm}}
  \caption{Repr\'esentation  2D d'un sous ensemble de la base $Eachmovie$ contenant 3 genres de films sur les deux premiers axes en utilisant la m\'ethode PCA: (a) donn\'ees dans l'espace d'origine.   (b) donn\'ees dans l'espace de redescription.}
  \label{fig:exemple1}
\end{figure}
A partir de cet ensemble de donn\'ees, nous avons construit un sous ensemble de jeu de donn\'ees contenant 75 films r\'epatis sur trois classes recouvrantes. La classe "Action" avec $21$ films, la classe "Com\'edie" avec $26$ films, la classe "Crime" avec $17$ films et $11$ films appartenant simultan\'ement \`a la classe "Action" et "Crime". Le processus de classification dans ce sous ensemble consiste \`a d\'eterminer le genre du film en se basant sur l'age, le sexe et la note d'\'evaluation des internautes. La figure 1.a montre la distribution des 3 genres de film sur les deux premiers axes en utilisant la m\'ethode PCA. Les objets repr\'esent\'es avec "$+$" sont des films recouvrants qui appartiennent au genre "Action" et au genre "Crime".
\begin{table}
\caption{Comparaison entre la m\'ethode OKM et la m\'ethode OKM-K sur le dataset $Eachmovie$.}
\label{tab:1} 
\begin{center}
\begin{tabular}{l|lll}
\hline \hline
{M\'ethode} & \footnotesize{Precision} & \footnotesize{Recall} & \footnotesize{{F-measure}}  \\
\hline
\footnotesize{OKM avec distance euclidienne    } & 0.557 & 0.788 & 0.616 \\
\footnotesize{OKM avec I-Divergence    } & 0.582 &0.687  &0.630   \\
\footnotesize{OKM-K avec noyau polynomial (d=0.25)     } & \textbf{0.700} &0.615  & 0.665  \\
\footnotesize{OKM-K avec noyau RBF ($\sigma=2$)     } & 0.628 & \textbf{0.851} & \textbf{0.721}  \\
\hline \hline
\end{tabular}
\end{center}
\end{table}
En projetant ces donn\'ees dans un espace infiniment dimensionnel en utilisant un noyau RBF de param\`etre $\sigma = 2$, nous remarquons une am\'elioration de la repr\'esentation des films recouvrants dans la figure 1.b puisque ces films recouvrants se trouvent \`a l'extr\'emit\'e des films de type "Action" et les films de type "Crime".

 Nous avons effectu\'e dix ex\'ecutions de chaque m\'ethode avec les m\^emes initialisations des clusters dans chaque ex\'ecution. Le tableau $1$. montre les diff\'erents r\'esultats obtenus. Nous remarquons que la m\'ethode OKM-K utilis\'ee avec un noyau RBF donne la valeur "F-measure" la plus \'elev\'ee. L'utilisation du noyau RBF a permis d'am\'eliorer simultan\'ement la mesure de pr\'ecision et la mesure de rappel par rapport \`a la distance euclidienne.
\section{Conclusion}
Nous avons propos\'e dans cet article la m\'ethode OKM-K qui permet explicitement de repr\'esenter les donn\'ees dans un espace de dimensionnalit\'e sup\'erieur \`a l'espace d'origine par l'utilisation de l'astuce de noyau. La recherche des recouvrements optimaux est effectu\'ee dans cet espace dimensionnel \`a travers la maximisation it\'erative d'une fonction objective. L'avantage de cette m\'ethode consiste en sa capacit\'e \`a identifier les clusters de formes non sph\'eriques. Les r\'esultats empiriques obtenus prouvent la performance de classification de la m\'ethode OKM-K par rapport \`a la m\'ethode OKM.

Comme travaux futurs, nous pr\'evoyons de profiter de l'utilisation des m\'ethodes \`a noyaux dans OKM-K pour appliquer la classification recouvrante sur des donn\'ees structur\'ees non vectorielles telles que les arbres et les histogrammes.

\smallskip


\medskip

\noindent {\large{\bf Bibliographie}}
\medskip

\small

\noindent [1] Banerjee, A., Krumpelman, C., Basu, S., Mooney, R. and Ghosh, J. (2005) Model based overlapping clustering. {\it In International Conference on Knowledge Discovery and Data Mining}, Chicago, USA. SciTePress.

\noindent [2] Camastra, F. and Verri, A. (2005) A novel kernel method for clustering. {\it IEEE Transactions on Pattern Analysis and Machine Intelligence}, 27:801-804.

\noindent [3] Cleuziou, G. (2007) Okm : une extension des k-moyennes pour la recherche de classes recouvrantes. {\it Revue des Nouvelles Technologies de l'Information}, Cpadus-Edition RNTI-E, 2:691-702.

\noindent [4] Cleuziou, G. (2008) An extended version of the k-means method for overlapping clustering. {\it In International Conference on Pattern Recognition ICPR}, pages 1-4, Florida, USA. IEEE.

\noindent [5] Deodhar, M. and Ghosh, J. (2006) Consensus clustering for detection of overlapping clusters in microarray data. {\it workshop on data mining in bioinformatics. In International Conference on data mining}, pages 104-108, Los Alamitos, CA, USA. IEEE Computer Society.

\noindent [6] Fellows, M. and  Guo, J., Komusiewicz, C., Niedermeier, R.,  Uhlmann, J.  (2009) Graph-Based Data Clustering with Overlaps. {\it Computing and Combinatorics}, 516--526,USA.

\noindent [7] Heller, K. and Ghahramani, Z. (2007) A nonparametric bayesian approach to modeling overlapping clusters. {\it 11th International Conference on Artifical Intelligence and Statistics}, San Juan, Puerto Rico.




\end{document}